# A Survey on Large Language Model-empowered Autonomous Driving

Yuxuan Zhu[a,1] , Shiyi Wang[b,1], Wenqing Zhong[c], Nianchen Shen[b], Yunqi Li[b], Siqi Wang[a], Zhiheng Li[b], Cathy Wu[d], Zhengbing He[d, *], Li Li[a, *]

[a] Department of Automation, Tsinghua University, Beijing, China
[b] Shenzhen International Graduate School, Tsinghua University, Shenzhen, China
[c] School of Vehicle and Mobility, Tsinghua University, Beijing, China
[d] Laboratory for Information & Decision Systems, Massachusetts Institute of Technology, Cambridge MA, United States

[*] Corresponding author. E-mail: hezb@mit.edu (Z. He) li-li@tsinghua.edu.cn (L. Li)
[1] The two authors contribute equally to this work.

**ABSTRACT** Artificial intelligence (AI) plays a crucial role in autonomous driving (AD) research, propelling its development towards intelligence and efficiency. Currently, the development of AD technology follows two main technical paths: modularization and end-to-end. Modularization decompose the driving task into modules such as perception, prediction, planning, and control, and train them separately. Due to the inconsistency of training objectives between modules, the integrated effect suffers from bias. End-to-end attempts to address this issue by utilizing a single model that directly maps from sensor data to control signals. This path has limited learning capabilities in a comprehensive set of features and struggles to handle unpredictable long-tail events and complex urban traffic scenarios. In the face of challenges encountered in both paths, many researchers believe that large language models (LLMs) with powerful reasoning capabilities and extensive knowledge understanding may be the solution, expecting LLMs to provide AD systems with deeper levels of understanding and decision-making capabilities. In light of the challenges faced by both paths, many researchers believe that LLMs, with their powerful reasoning abilities and extensive knowledge, could offer a solution. To understand if LLMs could enhance AD, this paper conducts a thorough analysis of the potential applications of LLMs in AD systems, including exploring their optimization strategies in both modular and end-to-end approaches, with a particular focus on how LLMs can tackle the problems and challenges present in current solutions. Furthermore, we discuss an important question: Can LLM-based artificial general intelligence (AGI) be a key to achieve high-level AD? We further analyze the potential limitations and challenges that LLMs may encounter in promoting the development of AD technology. This survey can provide a foundational reference for cross-disciplinary researchers in related fields and guide future research directions.

## TABLE OF CONTENTS

## 1. Introduction

Autonomous Driving (AD) has emerged as a pivotal research area within the realm of modern transportation whose recent development deeply relies on that of Artificial Intelligence (AI). The evolution of AI has consistently acted as a catalyst for the advancement of AD, and even the simplest Advanced Driver-Assistance Systems (ADAS) require AI within their implementation. Thus, the development of AD solutions can be comprehensively understood through the lens of AI design.

Two distinct ways of designing AI, namely, modularized and end-to-end solutions, shaped the two common solutions for AD as shown in Fig.1. The first solution, i.e., **modularized solution**, was a legacy of pre-AI system design. Such solutions break down AD into several independent tasks which typically include perception, prediction, planning, and control. While this modularization simplifies the implementation of individual tasks, it often struggles with system integration. Different models, each aimed at independent objectives, can lead to unavoidable gaps and conflicts within the system, resulting in sub-optimal performance. Enhancing coherence through implementation can be therefore a formidable challenge.

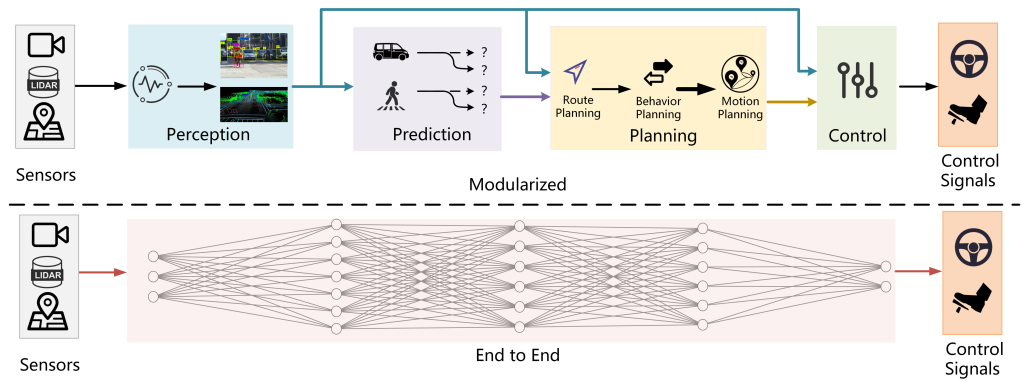

**Fig. 1.** Autonomous driving systems: modularization vs. end-to-end

The second solution, i.e., **end-to-end solution**, attempts to address these issues through a process that mimics human behavior. End-to-end solution utilizes large-scale neural networks and controls the vehicle directly based on sensor inputs. Different implementations, including imitation of human driving or direct training based on the control results, have been proposed. Nevertheless, all these methods along this path share common drawbacks of over-length information channels and complexity in network structures, resulting in difficulties of convergence during training as well as introduction of expert knowledge. Also, the data involved in the training of end-to-end algorithms is predominantly driving-related, while human drivers utilize common sense and other information in their driving process. These challenges limit the further improvement of end-to-end algorithms.

In addition to these specific issues, both solutions face some challenges in real-world applications, including ensuring robustness, validation, interpretability, and efficient human-vehicle interactions. Consequently, addressing these challenges has become a primary focus of AD research, highlighting the need for effective solutions.

Since the advent of ChatGPT in late 2022, a new revolution has been ignited in the field of AI. Owing to its vast scale, the volume of data, and the techniques involved in its training (e.g., learning from human feedback), Large Language Models (LLMs) have acquired capabilities in reasoning, data generation, and understanding human intentions, among others. These abilities have enabled LLMs to surpass previous models across a wide array of Natural Language Processing (NLP) tasks. Applications of LLMs in various fields such as Intelligent Transportation Systems are on the rise [1]. Specifically, the capabilities of LLMs offer innovative solutions to the aforementioned challenges within AD research. For instance, reasoning abilities could aid in understanding and appropriately responding to unseen corner cases, improving robustness. Generative capabilities could be utilized for test case generation. Enhanced comprehension of human intentions could help address interpretability issues and improve human-machine interactions.

Recently, LLMs have pioneered a new realm within AD research. There is a growing conviction among researchers that LLMs could illuminate fresh perspectives on traditional AD solutions. Moreover, there is a viewpoint that LLMs have further paved the path towards Artificial General Intelligence (AGI) which brings us back to the debate on achieving fully AD. Some experts argue that a large-scale, generalized, powerful intelligence is necessary while others believe that a specific intelligent agent with a smaller scale would be sufficient for AD tasks. Standing at this crossroads, this paper provides a systematic overview of the latest AD advances employing LLMs from the perspective of AD system implementation, focusing on the discussion of the following questions:

- What are the current challenges within AD research? Specifically, this is a set of specific challenges (i.e., Challenges I to X in the main text) corresponding to different tasks within AD implementation.
- Can LLMs provide superior solutions to these challenges and how?
- What might be the ultimate solution for AD? What should be the goal for the optimization of AD algorithms?

To this end, Sections 2 and 3 first revisit the fundamental problems across different roadmaps for implementing AD. Section 4 provides a brief review of LLM development, and Section 5 then categorizes studies that utilize LLMs in various aspects of AD implementation and analyze how LLMs enhance AD research and address existing issues. Finally, Sections 6 to 8 present comprehensive discussions on the applications and challenges of LLMs in AD. We hope this work will further advance the application of LLMs in AD research.

## 2 Autonomous Driving Solutions

### 2.1 Modularization

A modularized solution decomposes the AD system into distinct modules. Typically, these modules include perception, prediction, planning, and control. In this section, we go through all these modules, from their function to their development, and finally come to current challenges.

#### 2.1.1 Perception

Perception involves the collection of environmental information, extraction of useful knowledge, and formation of judgments to understand the environment. The accuracy and comprehensiveness of perception are vital for Autonomous

Vehicles (AVs) to navigate complex traffic scenarios effectively. Perception tasks were initially limited to the question of **"How to see?".** This referred to the recognition and tracking of surrounding objects, both static (e.g., lanes, traffic lights, and other traffic infrastructures) and dynamic (e.g., vehicles and road users). With the progress of sensing technologies and the development of machine learning (ML) over the last two decades [2,3], these fundamental tasks no longer pose challenges. Basic applications like lane detection or traffic signal recognition can now be executed correctly under most circumstances, paving the way for the promotion of low-level Advanced Driver-Assistance Systems (ADAS).

More advanced sensors and neural network structures have also shed new light on perception algorithms, which has been evolving from object-level to scenario-level. The emergence of Bird's Eye View (BEV) and Transformer-based methodologies has made the recognition of comprehensive scenarios possible. BEV projects multimodal 3D data around the vehicle onto a 2D map, ensuring consistent data representation. Transformers, originally devised for NLP, have demonstrated their efficacy in modeling multi-source heterogeneous data due to their robust attention mechanisms. This has empowered BEV representations to adeptly capture comprehensive spatial information. Approaches following this paradigm, such as BEVFormer [4], have showcased superior performance across various perception tasks, emerging as the predominant perception solution. However, BEV's deficiency in height information restricts its efficacy in representing 3D volumes. Occupancy networks have bolstered BEV by directly learning 3D information in voxels, portraying 3D surfaces as neural network decision boundaries, and obviating the need for LiDAR point clouds. They amalgamate geometry and semantics to precisely depict scenes, enhancing perception efficiency and accuracy.

With the obtain of scenario information no longer challenging, the current research focus has shifted to the ultimate goal of developing a comprehensive understanding of the environment through reliable and detailed representations, namely the question of **"What to see?".** This requires the perception system to nonspecifically recognize surrounding objects, identify their attributes and interactions, and understand the scenario thoroughly. Historically, AD perception algorithms have often amalgamated temporal and 3D spatial data into 2D object detection frameworks (e.g., YOLO [5], CenterNet [6]), amalgamating inputs from LiDAR , cameras, and leveraging diverse deep learning models such as PointNet [7] for information processing. Nevertheless, the integration of features across different scales (e.g., temporal, spatial, task-related) presents challenges.

Despite considerable advancements, existing perception solutions still face several challenges. First,

- **CHALLENGE I: How to improve the performance of perception systems in the real world or uncontrolled environments?**

Current learning-based methods are heavily dependent on training data, and their performance significantly deteriorates when encountering corner cases that exist within the long-tail distribution of real-world data. Second,

- **CHALLENGE II: How to form a comprehensive understanding of complex scenarios like humans did?**

Rather than understanding the scenario, current approaches to scene understanding are more like simply integrating all data and modalities. Last,

- **CHALLENGE III: How to enhance the efficiency of processing the vast amount of sensor data collected and establish a more unified data annotation method?**

These challenges underscore the complexity of AD and the need for ongoing research and innovation in the field.

### 2.1.2 Prediction

Prediction is a key component in the AD process, whose goal is to predict the upcoming motion trajectories of objects, mainly pedestrians [8] and vehicles, based on their past motion trajectories. The module wasn't initially part of the AD workflow. With the evolution of AD solutions in the last two decades, prediction gradually caught more attention as an independent part, bridging the gap between perception and planning. Functionally, it directly utilizes perception as an input, while its output serves as a crucial reference for subsequent planning and control tasks. From a temporal perspective, prediction represents a transition from the past to the present and future, a transition that is integral to an end-to-end workflow. Traditional prediction methods are predominately model-based. These include physical models, intention models, and interaction models. Such methods faced limitations in handling the uncertainty of trajectories, especially over longer time horizons.

In the last decade, learning-based methods have progressively dominated the solutions of prediction tasks [9,10]. For instance, Recurrent Neural Networks (RNNs) [11] and their derivative network architectures, such as Long Short-Term Memory (LSTM) [12] networks, are prevalently employed in deep learning-based prediction paradigms. Other utilized network architectures encompass Convolutional Neural Networks (CNNs) [13] and Graph Attention Transformer [14]. These learning-based approaches have significantly improved the reliability and accuracy of predictions across wider periods, and advancement in perception, such as BEV [15], enabled multi-target collaborative prediction. This signifies an evolution in the predictive modeling landscape, underscoring the importance of collaborative prediction strategies in achieving superior predictive outcomes. Depending on the target, the latest prediction methods are capable of providing an accurate prediction of trajectory that lasts for a few seconds even more than ten seconds. This would be adequate in most cases for the backend AD tasks.

Current research on prediction tasks aims to improve the accuracy within dynamic environments to enhance the safety and efficiency of AD, which requires a focus on more than just trajectories, but also the situation. Vehicles must understand and respond appropriately to the social dynamics or varying environments, and this can be concluded as another specific challenge:

- **CHALLENGE IV: How to realize comprehensive situation-aware predictions in complex scenarios?**

Tackling this challenge would be another step toward more accurate and advanced prediction methods.

### 2.1.3 Planning

Planning refers to the process where an AV sets a future driving route or trajectory based on the given traffic environment and the vehicle's situation [16,17]. Depending on the specific function and planning scope (spatial and temporal), planning can be generally divided into route planning, behavioral planning, and trajectory planning (also known as motion planning). Specifically, route planning outlines a road network level path for the vehicle, which is commonly referred to as "navigation". Behavioral planning provides decisions at important waypoints along the planned route. Trajectory planning generates a precise spatio-temporal trajectory connecting waypoints for the vehicle to follow.

Despite various targets and constraints, different planning tasks can be formulated in a similar way, therefore sharing similar methods. For instance, primitive planning methods originated from traditional search methods including A* [18], Rapidly-exploring Random Tree [19], etc. These are known as search-based planning methods. Optimization-based methods utilize optimization theories in the search for optimal trajectories [16]. Compared to search-based methods, these methods are more time-efficient in complicated scenarios.

Learning-based methods are also emerging in planning. For instance, Reinforcement Learning (RL) is widely utilized in planning tasks [20], and a planning task is typically formulated as a Markov Decision Process. Imitation Learning (IL) provides a different paradigm for learning-based planning. Other methods combine neural networks with existing planning methods [21] or use neural networks to directly generate planned trajectories [22], offering a real-time, online solution for planning tasks.

The current research gap for planning methods mainly lies in two aspects. First,

- **CHALLENGE V: How to improve the performance of planning methods when facing complex kinematic or scenery constraints?**

This requires the system to better integrate information from the frontend modules while handling inherited uncertainties. Second,

- **CHALLENGE VI: How to bind the planning tasks to form a more integrated hybrid planning?**

This would benefit the planning process in terms of robustness and better performance.

2.1.4 Control

The final step in the traditional modularization is control, which involves the vehicle driving along the preset planned trajectory (trajectory following). A basic target for such a process is to minimize the error between the target trajectory and the real vehicle trajectory. Other control targets include improving stability or riding comfort [23].

Various controllers and methodologies have been developed for the control process. Fundamental control methods, such as pure pursuit [24], primarily account for the vehicle's kinematic constraints. In contrast, other methods incorporate the vehicle's dynamic characteristics to achieve more precise control. Controllers, such as Model Predictive Control (MPC), are adept at managing more intricate scenarios. Given the inherent stability of the vehicle relative to external environments and the nature of control problems, learning-based methods are less frequently employed in control tasks. However, the emergence of hybrid controllers, such as learning-based MPCs [25] that amalgamate traditional and learning-based controllers, is noteworthy.

The major challenge within vehicle control lies in one single problem.

- **CHALLENGE VII: How can controllers adapt to various, comprehensive scenarios?**

Real-world scenarios range from extreme operating conditions [26] in which vehicles are reaching the threshold of stability to personalized control requirements. Adapting to various scenarios requires both better robustness of the controllers and the room for precise tuning. The control module also needs to coordinate with the front-end modules within the AD solution to achieve better performance.

## 2.2 End-to-end

Compared to the modularized solution, end-to-end AD utilizes a different roadmap. In a narrow sense, end-to-end AD tries to mimic the way humans drive their vehicles, in which raw sensor inputs are directly mapped into trajectory points or control commands for the vehicle using one large-scale neural network [27]. The first attempt at end-to-end AD, Autonomous Land Vehicle in a Neural Network [28], dates back to the 1980s. It tries to map the input from cameras and laser detectors directly to the vehicle's steering control.

With the advancement of ML methods, end-to-end AD has undergone a blooming emergence within the last decade. The most widely applied learning technique within end-to-end AD is IL [29–31]. IL is a supervised learning method, which can be further divided into behavior cloning and inverse optimal control. Another learning method involved in end-to-end AD is RL [32], and techniques including policy distillation [33] are applied to enhance the performance of the algorithm. In addition, the continuous accumulation of datasets, increasingly refined testing environments, and evaluation metrics have further accelerated the practical application of end-to-end AD. AD systems, such as the Tesla FSD system, which are trained using end-to-end methods, have already been applied in open environments.

As research on end-to-end solutions deepens, the focus is shifting toward the core ideas behind these solutions rather than their form. The integrated approach of end-to-end solutions provides a unified channel for transmitting various types of information and data. This minimizes the loss and compression of information during its transfer between different modules or subtasks, which is considered one of the advantages of end-to-end solutions. Applying this idea, many generalized "end-to-end" applications of module subtasks are emerging, especially perception and planning modules that are divided into multiple hierarchical sub-tasks. For example, an "end-to-end" perception module [34] takes the data of sensors as input, while providing integrated and complete scene perception output. These generalized end-to-end modules make the training and execution of subtasks smoother and more efficient.

Meanwhile, the form of end-to-end AD systems is also evolving. For example, the latest Unified End-to-End Autonomous Driving (UniAD) solution [35], known as "modular end-to-end AD", integrates three major tasks and six minor tasks of AD. Each module remains relatively independent in terms of network structure. During training, each module is pre-trained before the entire system is trained, ensuring consistent training objectives. This approach has demonstrated excellent performance in closed-loop simulation verification, proving that the essence of end-to-end solutions is not necessarily to complete all tasks with a single network.

Although end-to-end solutions are rapidly developing and have solved some of the existing problems in modularized AD systems, some other challenges have yet to arise. Compared to modularized AD, end-to-end systems utilize neural networks on a larger scale and almost completely rely on training data, both of which add to challenges during the training process. For instance,

- **CHALLENGE VIII: How to establish datasets, including selecting specific data and generating new cases, with better quality so as to support the training of end-to-end algorithms?**

Also,

- **CHALLENGE IX: How to improve the training efficiency (e.g., design better reward functions) of end-to-end solutions?**

Finally,

- **CHALLENGE X: how to improve the interpretability of end-to-end solutions?**

These challenges are closely related to the performance of end-to-end AD.

## 3 Development of Large Language Models

Modern language processing models originated from two paradigms: rule-based and statistical. Rule-based language models rely on manually defined grammar, semantics, and pragmatics rules to generate natural language using a set of handcrafted rules. The limitation of this approach lies in the need for a significant number of manually set rules, making it challenging to cover all scenarios in real language corpora. Statistical language models, on the other hand, depend on vocabulary statistical distributions in large corpora, such as n-gram models that predict text using n words in context. These models marked the beginning of rationalist approaches in language processing.

With the advancement of deep learning, researchers began using neural network models to learn the complex structures and semantic information of natural language. RNNs became a classical model framework for handling natural language and other time-series problems. By introducing a recurrent structure and weight parameters shared across time steps, RNNs can handle natural language text sequences of different lengths and capture hidden features. LSTM networks [36,37], an improvement over traditional RNNs, introduced memory cells, selectively retaining or forgetting information to effectively address issues like gradient vanishing or exploding. Some works building upon this foundation include the Gated Recurrent Unit (GRU) [38]. In 2017, the Transformer [39], relying on its powerful contextual processing ability, gradually started being applied to NLP tasks. Compared to previous statistical models, these neural network-based models share better approaches to handle representations and features within language, thereby showing better performances in specific NLP tasks.

With the advancement of network structures, a new training method for language models, pre-training, also contribute to further evolvement of language models. In 2018, Google introduced BERT (Bidirectional Encoder Representations from Transformers) [40], a pre-trained language model based on attention mechanisms. It adopts a two-stage strategy, initially undergoing unsupervised pre-training on a large-scale corpus and then fine-tuning with supervised learning on specific tasks to adapt to downstream tasks. The first generative pre-trained model, GPT-1 [41], was also released by OpenAI in 2018. Utilizing a network model based on the Transformer architecture, specifically with a decoder-only structure, GPT-1 achieved superior performance on specific tasks through generative pre-training and discriminative fine-tuning. GPT-2, an advanced version released in 2019, also adopted a structure with only stacked encoders. Differing from GPT-1, it used an unsupervised learning mode, predicting output based solely on input and task probability models. These models can be referred to as "pre-trained language models", and they have important implications for zero-shot learning and transfer learning, paving the toward LLMs.

LLMs, with parameters reaching a certain scale (e.g., several billion or hundred billion), have attracted widespread attention as an emerging AI technology in the last three years. They inherit their structure and training paradigm from pre-trained language models, and benefit from the scalability and emergent effects provided by their increased scales. LLMs not only achieve significant performance improvements in language related tasks but also demonstrate capabilities such as context learning and can be applied in other different scenarios. In 2020, GPT-3 [42] was introduced, following the single-direction language model training of GPT-2 but with a larger model and dataset. Experimental results showed that GPT-3 achieved superior performance in small-sample learning scenarios. In March 2023, OpenAI released GPT-4 [43], and in May of the same year, the technical report for GPT-4 was published. Compared to the past three versions of GPT, GPT-4 has made significant advancements in small-sample learning, logical reasoning, multimodal understanding, security, and support for longer contexts.

Similar LLMs also include Genimi, which leverages advanced language models like PaLM 2 to excel in NLP and information retrieval [44]; Claude, which emphasizes safety and interpretability and is particularly valuable in scenarios requiring secure and aligned AI systems [45]; LLaMA, an open-source model that has a significant impact on the research community by providing an accessible platform for exploring LLMs [46]; and Command R, which focuses on retrieval-augmented generation to enhance content creation and business applications [47].

## 4 Applications of Large Language Models in Autonomous Driving

In this section, we introduce existing works of applying LLMs in AD research by following the perspective of AD solutions, from the modularized algorithm to end-to-end implementations. Then, we would focus on how LLMs solve existing challenges in AD algorithms and provide our insights on it.

### 4.1 Modularization

#### 4.1.1 Perception

In AD perception tasks, LLMs have promoted the improvement of traditional perception task solutions and opened up new research directions. The current application of LLMs in perception tasks mainly focuses on the following aspects:

First, LLMs can be utilized directly in traditional perception tasks. Even though the language is difficult to use directly for acquiring or describing the surrounding environment, it can serve as a buffer between modalities and combine with them, such as vision and LiDAR, for simple inference. This enhances the efficiency of perception tasks and optimizes performance in different scenarios. For example, [48] used LLMs to analyze language cues of pedestrians and thereby improve the accuracy of pedestrian detection. [49] utilized LLMs in the process of analyzing LiDAR point cloud inputs, providing descriptions of the detected objects. [50] employed LLMs for generating prompts describing objects with specific labels. A reverse tracking workflow encodes multiple images from different angles and text input, predicting tracking through reasoning. [51] exploited Vision-Language Models (VLMs) in the segmentation of traffic scenarios under complex weather conditions. The module is capable of providing detailed descriptions of the scenario and more accurate segmentation under various conditions.

Moreover, thanks to LLMs' ability to understand complex scenarios, LLMs have been directly used for the comprehensive perception of complex scenarios, integrating various perception information to understand complex traffic environments and driving situations. LLMs improve the understanding of complex interactions within the AD system by classifying, recognizing, and understanding the relationships between dynamic and static elements in traffic scenes. They also reason about the interactions of these elements in time and space, providing a solution to **Challenge II**. For instance, [52] utilized a traditional 3D-perception pipeline to reinforce the perception ability of VLMs. On the road with GPT-4V(ision), [53] demonstrated the potential of VLMs in handling out-of-distribution scenarios and recognizing intent in actual driving environments. [54] used camera-only data as inputs, forming a semi-disentangled output representation of the environment which can be used in downstream AD tasks. [55] tested the ability of different LLMs to make decisions based on their understanding of the environment and traffic rules in both simulation and actual vehicles.

The comprehensive ability of LLMs also throws light on the issue of corner cases, which has been a concern (**Challenge I**) within current perception algorithms. [56] analyzed the abilities of LLMs in understanding AD scenarios and identified the key abilities of LLMs for achieving human-like performances, pointing out that LLMs are capable of solving the crucial problem of corner cases in AD through comprehension of the situation. [57] explored how LLMs can be applied to identify edge cases and present a monitoring framework for detecting semantic anomalies in vision-based descriptions. The framework is proven to function in different perception scenarios and outperforms traditional methods in terms of accuracy. [58] utilized LLMs in retrieving BEV features from complex driving scenarios, which was also proved effective when dealing with corner cases.

Lastly, LLMs also enhance existing perception data in various forms through language modality. These applications not only facilitate transparent interpretability between humans and machines but also facilitate LLMs to further carry out planning and control, providing a possible solution for **Challenge III**. For example, regarding object reference, [59] proposed a unified vision-language model, which can refer to objects described in human instructions and generate bounding boxes in picture frames. [60] proposed a context-aware visual grounding framework based on GPT-4. The framework can respond to natural language instructions, refer to the key objects mentioned, and provide an analysis of a human's current emotion. Regarding language modal annotation, [61] utilized natural language as an additional modal of BEV. A brief label and a background description were designed for each object, enhancing the performance of BEV regarding scene understanding and can provide suggestions on the following planning and decision-making. In addition, Visual Question Answering (VQA) and explanations are common ways for extracting information from visual inputs. Several studies utilize LLMs in VQA for perception-related tasks. Based on the OpenFlamingo VLM framework [62], a vision-language model designed to imbibe human-like abilities as a conversational driving assistant was trained using Grounded Chain of Thought instructions to align with AD tasks [63]. The trained model is capable of providing an understanding of the scenario. [64] used LLMs to provide descriptions and reasoning of the surrounding environment. Apart from these perception-specific applications, [65] and [66] used VQA to guide the reasoning of LLMs during AD tasks.

### 4.1.2 Prediction

Due to the stronger task capabilities of LLMs, prediction tasks have evolved in two ways. On one hand, predictions are often implemented using LLMs in collaboration with perception or planning. This would improve the overall efficiency of the system and provide the prediction part with more accurate information. [67] embedded natural language descriptions and rasterized image representations of traffic scenarios to realize the prediction of trajectories. [68] used LLMs to derive motion cues from past trajectories and combines it with traditional methods to provide better predictions of future trajectories. [69] utilized LLMs in trajectory prediction based on a comprehensive understanding of the traffic scenario. Relative information is obtained through multi-modal inputs, and prompts are designed to guide the reasoning process of LLMs. [70] proposed an LLM-like model, State Transformers, which is a scalable trajectory model that unifies motion prediction and planning in AD as sequence modeling tasks. The model showed promising performance and generalization capabilities, especially in complex real-world scenarios by effectively leveraging large-scale datasets and transformer-based architectures.

On the other hand, LLMs utilize their reasoning abilities for scenario prediction, providing a partial solution for **Challenge IV**. [71] proposed a method to incorporate high-resolution information into multimodal LLMs. The language modal was used to conduct reasoning on low-resolution video and provide priors and highlights for high-resolution video frames. The framework also provided suggestions for upcoming behaviors based on predicted risks and scenarios. [72] utilized LLMs in real-time forecasting of potential accidents, by combining with real-time image road analysis provided by a large multimodal model.

### 4.1.3 Planning

LLMs' ability in reasoning and common sense provided them abilities to generate all kinds of planning for vehicles and serve as the vehicle driver. Typically, the planning tasks include route planning, behavior planning, trajectory planning, and hybrid planning.

Regarding route planning, LLMs can use map information, traffic rules, and user intentions to plan the optimal route for vehicles. Factors such as traffic congestion, road construction, and weather conditions can be considered to choose the most convenient and safest path. [73] applied LLMs in verbal descriptions of surrounding environments, navigation instructions, and other related information to provide meta-action level navigation sequences in an urban environment. [74] used LLMs to extract the user's intention and provide route planning in the form of description and map annotation accordingly. [75] explored LLMs' applications in planning urban-delivery routes, revealing LLMs' ability in optimizing real-world routes based on learning human driver's behaviors and integrating domain-specific knowledge.

Regarding behavior planning**,** LLMs can decide the behavior of vehicles under different circumstances, such as accelerating, decelerating, changing lanes, and avoiding obstacles. Such kind of meta-action decisions can be understood and executed using natural languages, requiring a comprehensive consideration of vehicle dynamics, surrounding environment, and the behavior of other vehicles. [76] proposed a continuous learning framework to enhance the behavior decision-making of multimodal LLMs in AD. [77] utilized LLMs to provide meta-action level behaviors based on a specific situation. LLMs are capable of learning from the experience of expert drivers and improving their driving skills gradually. [78] proposed a knowledge-driven framework for LLMs to conduct reasoning based on the scenario and provide the optimal action. It can also reflect on its past behavior to gain experience for future reference. [79] proposed a novel decision-making framework to enhance commonsense reasoning and proactive decision-making capabilities. It also presented a cyber-physical feedback system for continuous learning and improvement, and a method for safe human-machine interaction through natural language understanding.

Regarding trajectory planning, beyond quantitative meta-actions, LLMs can further provide a more precise trajectory for the vehicle, such as trajectories include turning, overtaking, and parking. [80] utilized a closed-loop framework for LLMs to provide qualitative motion planning in complex scenarios. The Co-Pilot framework is capable of providing a detailed trajectory in the form of coordinate points for the vehicle to follow [81].

Regarding hybrid planning, LLMs' reasoning ability also makes hybrid planning (**Challenge V**) possible. Such a hybrid pipeline integrates different levels of planning and even part of perception to form an "end-to-end" planning solution. [52] proposed a Chain-of-Though (CoT) containing scene description, scene analysis, and hierarchal planning to form a trajectory plan for the vehicle. [82] proposed a "GPT driver" that utilizes GPT as the motion planner for AD tasks, implemented through a "Prompting-Reasoning-Fine-Tuning" process. [83] utilized LLMs in real-world driving tasks, in which LLMs are used for generating codes for planning vehicle motion. Downstream controllers were also utilized to control the vehicle. [84] integrated LLMs into AD systems, and LLMs served as a trajectory planner or behavioral planner, focusing on safety verifying of the system. [85] proposed a novel AD planning system based on Multimodal LLM, which integrates scene representation, CoT reasoning, and simulation-based reflection and evaluation to enhance planning safety. It demonstrated outstanding performance in handling both common and complex long-tailed scenarios with improved token efficiency in scene description.

Apart from direct applications of LLMs in planning, another common approach is to combine LLMs with existing planning methods. LLMs enhance the performance of traditional methods through reasoning or augmentation. This can be referred to as Model-Based Planning (solving **Challenge VI**). [86] proposed a framework that utilizes LLMs to enhance performance of perception, prediction, and planning. [87] aligned decisions from Multimodal LLMs with those from a behavioral planning module to enhance the performance of planning tasks.

### 4.1.4 Control

Due to the requirement for quick responses in control tasks, LLMs are currently difficult to directly replace existing controllers. However, thanks to the understanding and reasoning capabilities of LLMs, they can participate in control tasks at a higher level, such as combining with planning tasks, to improve efficiency and adaptivity to different scenarios (solving **Challenge VII**). [88] combined LLMs with MPC to precisely control the vehicle in an AD scenario. The LLM provides high-level decision-making in the process, and a corresponding matrix is used to finetune the MPC controller. The framework is also considered applicable for multi-vehicle control scenarios. The hybrid reasoning method proposed by [89] demonstrated better accuracy in calculating control signals for AD vehicles. [90] combined LLMs within an adaptive PID controller for truck platooning systems. [91] proposed a VLM and planning-based hybrid structure that combines decision-making and control signal generation.

Several attempts to further apply LLMs directly in the control of the AD vehicles were also made. Most of them combined planning and control, providing more precise controller information based on the meta-actions. [81] proposed a Co-Pilot Framework using ChatGPT as a controller selector, which can accomplish the desired tasks and adjust its output to meet human intention properly.

## 4.2 End-to-end

As a more systematical solution, LLMs have been involved in different steps within the implementation of end-to-end AD solutions as summarized below.

### 4.2.1 Data Fundamentals

Learning-based AD algorithms, particularly end-to-end solutions, greatly rely on massive data. Therefore, many new datasets used for training LLMs based end-to-end algorithms are proposed as part of the solutions. These datasets mostly contain natural language annotations, therefore providing a thorough channel through which language can be applied in both the training and the implementation process. Such annotations also serve as a kind of distilled knowledge and therefore improve the training efficiency. [92] used a workflow based on Graph VQA for implementing end-to-end AD, providing an overall solution including dataset, task, baseline, and metrics. [65] integrated video frames and texts as input, and the output of the model includes response and predicted control signals. The training process contains two stages, pretraining and mix-finetune, with the latter stage using ChatGPT annotated data. [93] introduced a close-loop end-to-end framework that integrates LLMs with multimodal sensor data to enable vehicles to understand and follow natural language instructions in complex driving scenarios. It also presents a new benchmark for evaluating language-guided closed-loop AD systems.

More representative LLM-related datasets are listed Table 1. It is worth noting that the contributions of those datasets are not limited to end-to-end solutions, as they provide essential references for the development and validation of all AD tasks. The increased proportion of natural language in these datasets, which transit from simple labels to natural language descriptions or question-and-answer formats, also contribute to perception related **Challenges II and III**. For instance, [50] proposed a dataset containing natural language annotations of objects in 3D driving scenarios based on the NuScene Dataset [94].

Apart from real-world scenario-based datasets, scene generations are also becoming an important aspect of AD datasets with the help of LLMs' generative abilities. Such ability helps solving the long-tail problem of data distribution and improve training efficiency. [95] proposed a framework, namely LCTGen, to generate dynamic traffic scenes for simulation based on description and map data. LLMs were used to convert a natural language description of the scenario to a structuralized representation of the scenario. [96] used GPT4 to generate codes that can be executed for generating CARLA scenarios based on dialogues. [97] proposed SimCopilot, which utilizes LLMs to convert natural language descriptions of object interactions into programming code for virtual road scenes, significantly streamlining the creation of interaction data for AD systems. [76] utilized SUMO and Carla to generate scenarios and multimodal prompts for LLMs to understand it. [98] employed LLMs and collaborative agents to enable editable, photo-realistic 3D driving scene simulations, addressing key challenges in user interaction, rendering quality, and asset integration for AD applications. [99] introduced an LLM-enhanced framework capable of generating diverse, high-quality driving videos from user descriptions, which can be utilized in the training of perception algorithms. The WEDGE dataset [100] directly uses the DALLE large model to generate scene images, simulating 2D road and traffic scenes under various weather conditions, which is a novel attempt to build an AD dataset.

**Table 1** LLM Related AD Datasets.

| Type | Dataset Name | Data Origin | Language Function | Scale |
|------|-------------|-------------|-------------------|-------|
| LLM Assisted Annotation | Reason2Drive [101] | Multiple | Question Answering (QA) on Perception and Reasoning | 600K Scenes |
| | NuPrompt [50] | NuScenes (Images Only) [94] | Description of objects | 35.4K Prompts |
| | DriveGPT4 [65] | BDD-X [102] | QA on Perception and Reasoning | 56K Scenes |
| | LingoQA [66] | Real World | QA on Perception | 419.9K QA pairs |
| MLLM Generated | WEDGE [100] | DALL-E Generated | Data Generation | 3360 Scenes |
| LLM Training Related | DriveLM-nuScenes [92] | NuScenes [94] | QA & Description of AD tasks | 0.44M QA pairs (Captions) on 4.8k Scenes |
| | DriveLM-Carla [92] | Carla | QA & Description of AD tasks | 3.75M QA pairs (Captions) on 0.18M Scenes |
| | NuScenes-QA [103] | NuScenes [94] | QA on Perception | 4.59M QA pairs on 34.1K Scenes |
| | Rank2Tell [104] | Real World | QA on Importance Classification | 116 Scenes, 20s each |
| | LaMPilot [105] | HighwayEnv [106] | Driving Instructions | 4.9K Scenes |
| | MAPLM [107] | THMA [108] | Description of objects and maps | 2M Scenes with annotation |
| | LMDrive [93] | Carla | Driving Instructions | 64K Scenes |

## 4.2.2 Large Language Models as End-to-end Agents

LLMs themselves can serve as agents, accomplishing all the driving tasks at the same time. [109] proposed Agent-Driver, which utilizes LLMs as an intelligent agent of the vehicle. The agent is designed to access information for the driving via function calls and act as a human. [110] proposed an architecture that encodes scenario information as numeric vectors and applied a pre-trained LLM to understand the scenario and provide motion-level control. It is also capable of providing the reasons for its actions, improving the interpretability of the solution. [111] presented a pioneering general world model for AD that integrates Multimodal LLMs and diffusion techniques, enabling it to directly predict control signals and generate future frames based on historical vision-action pairs, thus simulating an infinite driving experience. [112] implemented a VLM-based end-to-end AD pipeline through Graph VQA. In this approach, the visual questions guide the reasoning process through different stages, achieving remarkable performance in complex scenarios.

## 4.2.3 Training and Reflection

As describe in **Challenge IX**, the training of end-to-end AD solutions has always been an important topic. LLMs have enhanced the efficiency of such a process, mainly through better reasoning and reflections. [113] proposed a novel dataset and benchmark for end-to-end AD that includes sensor data, control decisions, and CoT labels to indicate the reasoning process. It also proposed a baseline model, DriveCoTAgent, trained on this dataset to generate CoT predictions and final decisions, demonstrating strong performance in both open-loop and closed-loop evaluations and enhancing the interpretability and controllability of end-to-end driving systems. DriveGPT4 is able to perform zero-shot generalization in a simulated environment, showing potentials in improving training efficiency [65].

**4.3 Other Aspects**

In addition to the direct implementation of AD tasks, LLMs are also widely applied in supporting applications for AD, including human-machine interaction, issues of safety, ethics, and fairness. Regarding human-machine interaction, [114] proposed a framework that can optimize the code for conducting AD tasks using LLMs through evaluation and iterations. [115] proposed a human-centric framework that integrates LLMs in the planning of AD, providing useful assistance in complex driving tasks and responding to different queries through reasoning. AccidentGPT can respond to a driver's query or automatically provide specific suggestions (e.g., slow down) and notices upon driving safety [116]. [117] tested the ability of different LLMs to understand commands from a human-centric perspective. [105] leveraged LLMs to translate human instructions into executable driving policies for autonomous vehicles in various scenarios. Regarding safety, ethics, and fairness, [118] discussed the potential ethical issue in the decision-making of LLMs corresponding to AD scenarios. [119] proposed a mechanism based on LLMs that can help human drivers or AD vehicles adapt trajectory planning to local traffic rules.

**5 Will Large Language Models be a Panacea to Autonomous Driving?**

In Section 4, we have systematically demonstrated the growing adoption of LLMs in AD. These applications span the entire spectrum of AD, with many showcasing the potential of LLMs to surpass existing AD algorithms and address the challenges previously discussed. Based on our survey and analysis, we tried to offer insights into how these existing challenges have been, or are anticipated to be, resolved through the progression of LLM-based AD solutions. Therefore, we categorize these insights into the following three levels, and describe the challenges in Table 2.

- **Solution Insight A**: LLMs have demonstrated significant capability in solving the corresponding challenge, and comprehensive solution based on LLMs can be expected.

- **Solution Insight B**: LLMs have demonstrated capability in solving the corresponding challenge, but the challenge may not be fully solved given current drawbacks of LLMs.

- **Solution Insight C**: LLMs can improve performance in related tasks, but might not be able to solve the key problems within the challenges.

As shown in Table 2, we can conclude that **LLMs have demonstrated strong potentials in providing effective solutions toward current challenges in existing AD algorithms**. Specifically, we believe that LLMs' performance in AD tasks primarily stems from the following aspects.

- **Common Sense.** Common sense is the fundamental level of practical judgment or basic factual knowledge that every adult is expected to have. It is a human's distilled understanding based on accumulated experiences and knowledge. Common sense plays a pivotal role in daily life. The ability of humans to rapidly acquire specific skills through imitations is closely related to it. For instance, an adult requires approximately 20 hours of training to pass a driving test. This is because the training primarily focuses on the operational skills of driving. Many other driving-related skills, such as recognizing traffic signals or signs, are intuitively understood and do not necessitate extensive learning. However, for AD models, each of these elements must be individually designed and trained. LLMs amass a form of "common sense" from extensive corpora. This can be viewed as the representation of specific semantic indicators in a high-dimensional space. For example, the representation vector corresponding to the concept of a "red traffic light" and "stop" may have a close projection on a dimension representing AD behavior. This enables LLMs to execute complex AD tasks with minimal prior instruction and makes few-shot learning possible.

- **Reasoning Capability.** Compared to former language models, LLMs are able to comprehend causal and logical relationships within the text inputs. Therefore, some researchers believed that LLMs are capable of reasoning just as human did. Such reasoning capability enhances the LLMs' understanding of the traffic scenarios, which is vital for AD tasks that necessitate forecasting future situations and making decisions accordingly. Furthermore, the reasoning capability of LLMs offers a potential solution to the "long-tail" problems encountered in AD. Even in corner-case scenarios, these models can make accurate decisions based on their reasoning.

- **Communication ability.** Another important ability of LLMs is that they can communicate fluently with humans. Revisiting human driving behavior, we can notice that language is predominantly used for navigation and route descriptions. LLMs can understand human intentions and provide appropriate outputs through reasoning, and therefore can directly participate in the driving process. In addition to the driving task itself, the ability to communicate with humans also benefits LLMs in the training and tuning process. The mutual understanding and interaction between humans and LLMs have partially solved the problem of neural networks as black box models.

**Table 2** LLM and Existing AD Challenges

| Challenge | Content | Representative Paper(s) | Solution Insights |
|---|---|---|---|
| I | Improve the performance of perception systems in real-world environments | GPT-4V [53], drive like a human [56], etc. | A |
| II | Form a human-like understanding of complex scenarios | Drive-VLM-Dual [52], etc. | B |
| III | Enhance the efficiency of data processing and establish a more unified data annotation method | Talk2Bev [61], Dolphins [63], etc. | A |
| IV | Realize comprehensive situation-aware predictions in complex scenarios | HiLM-D [71], etc. | A |
| V | Improve the performance of planning methods when facing complex constraints | Drive-VLM [52] etc. | B |
| VI | Integrate different planning tasks | DriveMLM [87], etc. | A |
| VII | Adapt to different control requirements | LanguageMPC [88], etc. | C |
| VIII | Improve training data quality of end-to-end solutions | LMDrive [93] etc. | A |
| IX | Improve training efficiency of end-to-end solutions | DriveLM [92] etc. | B |
| X | Improve the interpretability of end-to-end solutions | DriveGPT4 [65] etc. | B |

We noticed that these advantages of LLMs have partially solved several long-lasting drawbacks inherited from data-driven learning algorithms. These algorithms have been widely applied in AD implementations, leading to specific challenges in AD tasks. LLMs are therefore capable of providing solutions to them. Furthermore, we can conclude that signify a shift from data-centric models to a hybrid model that leverages both data and knowledge. This dual-driven approach combines the advantages of both paradigms. From the perspective of executing driving tasks, this shift also renders the implementation process more akin to human-like decision-making, a possible ultimate goal for AD. We anticipate a future where data and knowledge will coexist for a significant period, potentially embodied in the form of LLMs, with the role of knowledge gradually gaining prominence in AD solutions. Future research should focus on how to better integrate data-driven and knowledge-based methods to enhance the efficiency of training and implementation of AD solutions.

## 6 Limitations

Despite various advancements, we must point out that the further application of LLMs in AD faces many limitations and challenges. As a safety-critical scenario, these limitations desire more attention in future research. Some of these limitations are mainly due to drawbacks of current LLMs' performance, which inherit from their model structures, training method, or implementations.

- **The "hallucination" problem of LLMs.** "Hallucination" refers to the situation where LLMs give results that do not match the facts or user requirements without sufficient basis. This phenomenon is common in LLMs, and as a system with high safety requirements, the tolerance for such problems in the process of AD is very low. Therefore, measures must be taken to guard against hallucination problems. Existing research has shown that the hallucinations of LLMs mainly come from data, training, and inference processes. To address these causes, methods such as Retrieval-Augmented Generation (RAG), improving pre-training and tuning processes, and designing CoT can be used to improve. In addition, to prevent possible erroneous results from affecting the actual operation of the vehicle, insurance mechanisms can also be designed, and other rules can be used to judge the rationality of the output of LLMs.

- **Model response time.** Poor real-time performance is currently one of the shortcomings of LLMs. Whether the model is online or offline, the immense volume of LLMs makes their response delay hard to ignore. This could have serious consequences in the context of AD, especially tasks (like control) which are highly sensitive to response time. Solving this problem may be achieved by improving computing power on the one hand, and on the other hand, before real-time performance is resolved, LLMs may be more suitable for tasks with higher delay tolerance.

- **Lack of understanding of the physical world.** While being powerful in processing and generating text, LLMs have a significant drawback when it comes to understanding the physical world. They lack the ability to interact with the environment and learn from it, meaning they do not have a direct understanding of physical concepts such as gravity, momentum, or the texture of objects. They can't experience the world in the way humans or even some robots can. Their knowledge is based entirely on the text they were trained on, and they can hardly update that knowledge based on real-world experiences or sensory input. This limits their ability to accurately model or predict physical phenomena, and can lead to outputs that are nonsensical or incorrect in the context of the physical world. This is a significant limitation when applying these models to tasks that require a deep understanding of the physical world like driving a vehicle. It is important to further improve the ability of LLMs to effectively capture and understand real-world information [120,121].

Other challenges come from the combination of LLMs and AD tasks.

- **Privacy and Security considerations.** The data used by AVs is often highly sensitive, including details about specific locations, driving habits, and more. Such data is essential for finetuning LLMs for AD tasks. This raises significant data privacy and security concerns. For instance, if an LLM is trained on publicly available data that inadvertently contains personal information, it could potentially learn and reproduce this information, leading to privacy breaches. Ensuring this data is securely handled and that the models do not unintentionally leak this information is a critical challenge.

- **Bias caused by language**. We noticed that almost all LLMs now use English as the dialogue language, and experiments and research involving other languages are quite few. Since language has a stronger regional nature, we believe this could become a potential bias, that is, the performance of the model cannot be aligned when using different languages. This bias may be more obvious than algorithms based on visual and other modalities, and may also bring potential ethical and moral risks. In addition, issues such as training data privacy and dataset security are also worth paying attention to. We believe that further studies can be conducted to solve these issues.

These challenges need to be tackled before LLMs can be applied in real-world AD applications, but we believe that the rapid evolution of LLMs and related AD solutions would continuously provide new insights on these challenges.

# 7 Perspectives

Since the concept of AD was proposed, researchers have been exploring different paths to achieve this goal. There have been many discussions corresponding to different technical paths. Therefore, we would like to review some of these discussions and provide some of our perspectives about the ultimate solution for AD.

## 7.1 End-to-end or Modularized Autonomous Driving

The modular and end-to-end approaches have consistently been at the forefront of discussions on AD technology. The introduction section of this article summarized some of the benefits and drawbacks of these two paths. However, LLMs throw new lights to this discussion. On one hand, the emergence of LLMs has blurred the boundaries between modular and end-to-end approaches. The versatility of LLMs allows them to accomplish multiple tasks simultaneously, thereby dissolving the traditional modular boundaries. For example, many planning tasks executed by LLMs are directly based on raw sensor inputs. Functionally, such implementation cover everything from perception to planning, and in form, they can be considered close to an end-to-end implementation. We believe that as the generalization ability of the model strengthens, this blurring may become a trend.

On the other hand, researchers have started to pay more attention to the core of the end-to-end approach rather than the form itself. The advantage of the end-to-end solution can be summarized as providing a uniformed channel, reducing the loss of information transfer between different modules. In other words, as long as the complete transfer of information is ensured, the difference in form is no longer essential. This is also the origin of "modular end-to-end" of the UniAD [35]. Such kind of shift in the form of end-to-end approach may also provide solutions to existing problems such as testing and validation of end-to-end algorithms.

Therefore, we can believe that the distinction between end-to-end and modular in form will continue to be weakened, but considering the safety and robustness of the system, some mature modules (such as ADAS) will be preserved in practical applications for a long time.

7.2 Artificial General Intelligence and Driving Intelligence

Finally, we arrive at a long-standing debate in the field of AD: Is a highly advanced AGI indispensable for achieving optimal AD? On one hand, as we have mentioned earlier, the common sense and other knowledge that LLMs own have played a significant role in performing AD tasks. Although we cannot yet determine whether LLMs are an essential waypoint to AGI, they do meet people's expectations for AGIs to some extent. The capability of applying natural language enables them to efficiently learn from the vast corpus of human language and interact with humans in an easily understandable manner. On the other hand, driving skills for humans are relatively independent. For instance, an experienced truck driver may not have received higher education, while a researcher in AD might not possess a driver's license. This implies that general AI may not necessarily be the sole solution for optimal AD.

From an idealistic perspective, it seems easier to build a driving intelligence entity. We are still quite far from AGI, while driving intelligence is easier to implement with the maturation of large models, world models, etc. If we can develop algorithms specifically for driving intelligence, we might be able to address more issues associated with large-scale models. However, achieving this goal also pose many challenges. First, the definition of optimal AD is still somewhat vague. What kind of driving strategy can be called optimal is still a topic worth further research and discussions. In addition, there are some challenges in the implementation of the idealized optimal driving model itself. For example, due to the limitations of human cognition, it's challenging to precisely define what knowledge is needed for optimal driving. Figure 2 illustrates this from a knowledge perspective. Some knowledge required for optimal driving might not yet have a method to be summarized, such as the intuitive judgments made by experienced drivers in critical situations.

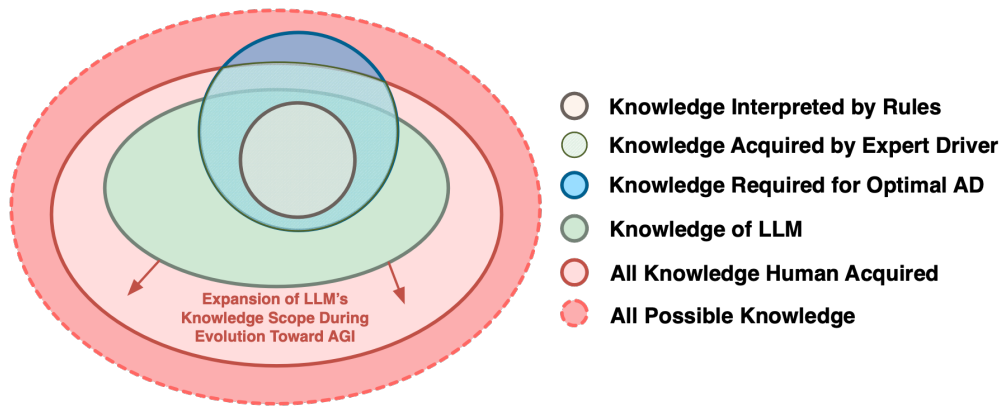

**Fig. 2.** Illustration on Knowledge Scope

On the other hand, we believed that LLMs will remain as one of the optimal forms of AGI-like agents in the near future. Therefore, LLM-powered AD will still be a notable research frontier. To further optimize the application of LLMs in AD tasks, we believe research can be conducted in the following areas. First, optimizing the structure of LLMs themselves and designing more efficient training methods. These improvements can enhance the model's capabilities in reasoning, common sense, etc. compared to existing models. Additionally, better designed structures can help reduce computational power consumption during fine-tuning and local deployment. This aids in deploying LLMs in intelligent vehicles, improving the issue of long response times, and reducing the cost of applying LLMs. Furthermore, various optimizations can be made in the integration of AD and LLMs. For example, introducing more AD-related data during the pre-training stage. The software and hardware structures of existing intelligent vehicles can also be optimized to support system-level integrated applications of LLMs [122].

In general, this issue may largely depend on the subsequent development of AI technology: whether the development of general AI can achieve rapid breakthroughs, or whether driving intelligence models can be implemented faster. We believe that for a considerable period of time, both approaches have their strengths and will coexist or serve as backups for each other, just like modular and end-to-end solutions.